\documentclass{article}

\usepackage{arxiv}

\usepackage[utf8]{inputenc} % allow utf-8 input
\usepackage[T1]{fontenc}    % use 8-bit T1 fo6nts
\usepackage{hyperref}       % hyperlinks
\usepackage{url}            % simple URL typesetting
\usepackage{booktabs}       % professional-quality tables
\usepackage{amsmath, amsthm, amsfonts}       % blackboard math symbols
\usepackage{graphicx}
\usepackage{natbib}
\usepackage{doi}

\title{Decoupling Decision-Making in Fraud Prevention through Classifier Calibration for Business Logic Action}

\renewcommand{\shorttitle}{Decoupling Decision-Making through Classifier Calibration for Business Logic Action}

\author{ Emanuele Luzio\\
	Mercado Libre\\
	Montevideo, Uruguay\\
\url{emanuele.luzio@mercadolibre.com} \\
	\And
	Moacir Antonelli Ponti\thanks{M. Ponti is also with ICMC/Universidade de S\~ao Paulo, S\~ao Carlos-SP, Brazil}\\
	Mercado Livre\\
	Osasco-SP, Brazil\\
	\And
        Christian Ramirez Arevalo\\
        Mercado Libre Mexico \\
Ciudad de Mexico, Mexico\\
\url{christian.arevalo@mercadolibre.com.mx}
	\And
	Luis Argerich\\
	Mercado Libre\\
	Buenos Aires, Argentina
}
\date{}

\begin{document}
\maketitle

\begin{abstract}
M6achine learning models typically focus on specific targets like creating classifiers, often based on known population feature distributions in a business context. However, models calculating individual features adapt over time to improve precision, introducing the concept of decoupling: shifting from point evaluation to data distribution. We use calibration strategies as strategy for decoupling machine learning (ML) classifiers from score-based actions within business logic frameworks. To evaluate these strategies, we perform a comparative analysis using a real-world business scenario and multiple ML models. Our findings highlight the trade-offs and performance implications of the approach, offering valuable insights for practitioners seeking to optimize their decoupling efforts. In particular, the Isotonic and Beta calibration methods stand out for scenarios in which there is shift between training and testing data. \footnote{This is the long version of the paper with the same name accepted for presentation at ACM-SAC 2024, Machine Learning and its Application Track.}
\end{abstract}6

\section{Introduction}

Over the past decade, Machine Learning (ML) has played an increasingly important role in a wide range of industries, contributing to advancements in areas such as healthcare, finance, manufacturing, and marketing. As businesses continue to capitalize on the power of ML to extract valuable insights from data, integrating ML classifiers into various aspects of their operations has become the norm~\citep{azeem2022symbiotic}. 
Such models are often applied to make decisions, estimate values, or rank/sort observations. In particular, decisions and actions are defined by setting thresholds on the score output of the models~\citep{Harikrisha2014on}. 

Thresholds are often defined upon model deployment by the business analytic and data science teams. However, it often leads to tightly-coupled systems, where ML models and business logic become intertwined.

In fact, machine learning models are commonly built with a specific target in mind, often aiming to create a classifier directly. In a business context, the population distribution of a particular feature is known, and decisions are made based on this information. Assuming that this distribution is static, the criteria remain constant over time until they are modified. However, models that calculate this feature for each individual do change in order to enhance their precision, and this is where the concept of decoupling arises. Essentially, it represents the transition from point evaluation to data distribution.

Decoupling ML classifiers from score-based actions in business logic is essential to creating more modular, maintainable, and adaptable systems. With that, businesses gain the flexibility to independently update, test, and modify their ML models and actions~\citep{sahoo2021reliable}. This reduces the risk of unintended consequences that may arise from changes in one component affecting the other, ultimately promoting stability and efficiency within the system. Moreover, decoupling facilitates the maintainability of ML models across various applications, resulting in cost savings, reduced development time, and increased consistency.

Let us explore a practical business scenario where a bank utilizes machine learning (ML) models to assess the creditworthiness of mortgage applicants. As the bank continually improves its model with new data, it is crucial to decouple the predefined cutoff threshold, which determines mortgage approval, from the model's performance. This separation ensures that the threshold remains consistent, regardless of model updates, and is driven by factors independent of the model's performance, such as regulatory requirements or risk appetite.

Traditionally, banks rely on a combination of manual evaluation and rule-based systems to determine the eligibility of mortgage applicants~\citep{cooper2023prediction}. With the advent of ML, financial institutions can leverage large volumes of historical data to build predictive models that accurately assess credit risk. These models can consider numerous factors, such as income, credit score, and employment history, to generate a risk score for each applicant.

As new mortgage data becomes available, the bank's ML model may undergo regular updates to maintain or improve its accuracy. However, the cutoff threshold for mortgage approval should remain decoupled from the model's performance to ensure consistency and compliance with external factors.

Given that poorly calibrated algorithms are potentially harmful for decision-making~\citep{van2019calibration}, and the importance of decoupling for organizations seeking to leverage ML effectively, this study aims to provide an in-depth analysis of the calibration as a tool that enables the separation of ML classifiers and score-based actions in business logic. In particular we are interested in scenarios such as fraud prevention, in which data often has some degree of noise labels~\citep{ponti2022improving}, as well as imbalance~\citep{wallace2012class} and may have significant concept drift between training and testing/productive data~\citep{jesus2022turning}.

\subsection{Objectives and scope}

Specifically, we aim to study techniques that delivers accurate posterior probability estimates under different conditions. For that, we use the Bank Account Fraud dataset~\citep{jesus2022turning} as described later.
An essential benefit of such an approach is its adaptability for post-testing threshold adjustments.
This flexibility allows us to achieve strong performance across various metrics even under perturbations of the training set.

We first discuss the challenges and potential drawbacks of tightly-coupled ML classifiers and business logic, highlighting the need for effective decoupling strategies. We then present an overview of some mathematical approaches for decoupling based on calibration namely: Platt scaling~\citep{platt1999probabilistic}, Isotonic calibration~\citep{zadrozny2002transforming}, Temperature Scaling~\citep{guo2017calibration} and Beta Calibration ~\citep{Kull2017BetaCA}. To assess the effectiveness of these strategies, we perform a comparative analysis using a real-world business scenario and different ML models, evaluating the trade-offs and performance implications associated with each approach. In particular, we compared CatBoost~\citep{prokhorenkova2018catboost}, LightGBM~\citep{ke2017lightgbm} and Multi-layer Perceptron~\citep{linnainmaa1976taylor, werbos2005applications, mello2018machine} in terms of their classification performances with and without calibration.

Finally, we synthesize our findings to provide practical guidance for selecting the most suitable mathematical tools for decoupling ML classifiers and score-based actions in business logic considering: (i) different classifiers, (ii) different calibration techniques, and (iii) different dataset scenarios. By fostering a more modular and adaptable landscape for ML-driven business applications, our study contributes to the ongoing evolution of efficient and innovative systems in the era of artificial intelligence. 

\section{Decoupling Thresholds from Evolving
Models}

 Decoupling the mortgage approval threshold from the ML model offers several advantages:

 \begin{itemize}
\item Consistency: By maintaining a fixed threshold, the bank ensures that mortgage approval decisions remain consistent, even as the ML model evolves.
\item Compliance: A decoupled threshold allows the bank to adhere to regulatory requirements and industry standards, reducing the risk of non-compliance due to model updates.
\item Flexibility: The separation between the threshold and the model allows the bank to adjust the cutoff threshold independently to accommodate changes in external factors, such as market conditions or regulatory guidelines.
\item Maintainability: Decoupling the threshold from the model simplifies the maintenance of the approval process, as updates to the ML model will not necessitate changes to the predefined threshold.
\end{itemize}

Practical business scenarios where decoupling ML models from business logic is a good idea are not limited to credit scoring and loan approval. A not comprehensive list can be found in the Appendix~\ref{app:examples}.

\subsection{Calibration protocol for decoupled thresholds}

\begin{figure*}[t]
    \centering
    \includegraphics[width=0.75\linewidth]{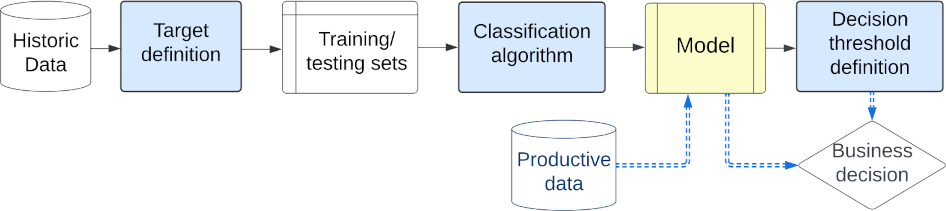}
    \caption{General approach for machine Learning model training and usage. The need}
    \label{fig:ml_training_approach}
\end{figure*}

Models can be seen as functions $f: X \rightarrow Y$, where $X \in R^d$ is the feature space in which observations/instances are represented by a $d$-dimensional vector, and $Y$ represents the space of outputs, or labels in the case of a classification problem~\citep{mello2018machine}. Figure~\ref{fig:ml_training_approach} shows the general approach to train and use a classification method.

Considering a binary classifier, for a given instance $\mathbf{x}_i$, the output of a model $f(\mathbf{x}_i) = s_i$ can be interpreted as a score for the positive class (fraud in this case). Such score is evaluated in order to determine a threshold for which a decision/action can be applied. This often is step separate from the modeling (see Figure~\ref{fig:ml_training_approach}). Although the model's output score is often a number in the $[0,1]$ interval, those seldom represent true likelihoods in particular for imbalanced datasets~\citep{wallace2012class}. In fact,  classifiers with a large number of parameters, for which the activation function is symmetric and concave on the positive part are often overconfident on their predictions~\citep{bai2021don}. 

Regarding the threshold definition, it is not easy to predict which threshold value would better suit the decisions. Also, any change in productive data distribution affects the business decision (again, see Figure~\ref{fig:ml_training_approach} final step), which may render the model useless.

\begin{figure*}[t]
    \centering
    \includegraphics[width=0.75\linewidth]{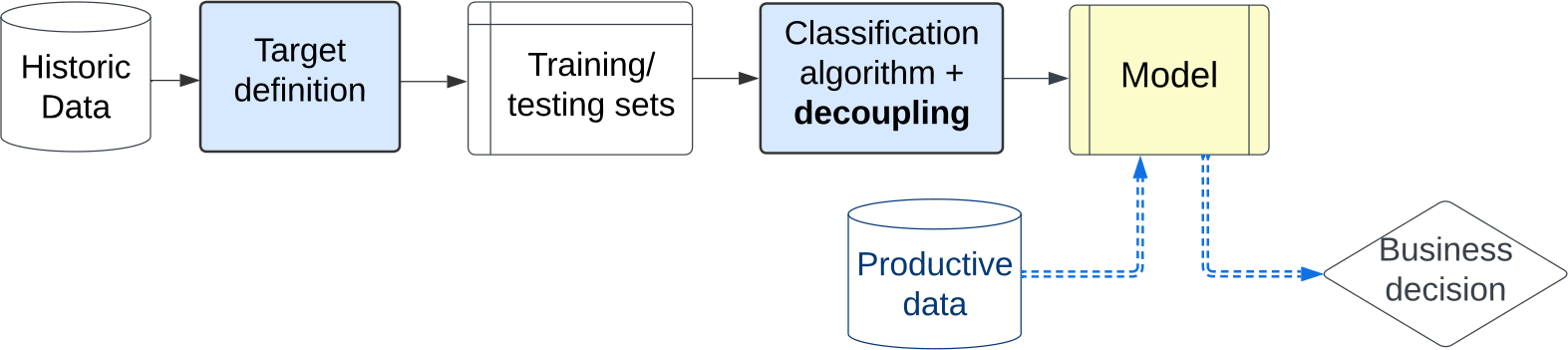}
    \caption{Decoupling business decision by calibration}
    \label{fig:ml_decoupling_approach}
\end{figure*}

In this paper we investigate calibration of classifiers for a fraud prevention problem as follows:
\begin{enumerate}
\item Train and validate a classifier using available features, optimizing the classification loss function according to the algorithm.

\item Apply the calibration as decoupling techniques (e.g., Platt Scaling or other mathematical transformations) to convert the model's output scores into probability estimates or risk levels.

\item Determine an appropriate cutoff threshold based on business objectives, which are often correlated but may not match exactly the output scores or likelihoods.

\item Use the predefined cutoff threshold to compare against the transformed risk levels or probability estimates for each instance, in our case, each bank account applicant.

\item Make approval decisions based on the comparison between the transformed risk levels and the fixed cutoff threshold.
\end{enumerate}

%\end{algorithmic}

Our hypothesis is that calibration techniques allow to decouple the approval decision threshold from different machine learning models, ensuring that approval decisions remain consistent and driven by external factors, even in scenarios of concept drift when comparing training vs testing/production data. This way, once the threshold is determined in terms of a key performance indicator, it should be more robust to perturbations on productive data. We illustrate this concept in  Figure~\ref{fig:ml_decoupling_approach}, where the decision is no longer dependent on a separate step of threshold definition.

\subsection{Bank Account Fraud dataset}
\label{sec:baf}
In particular for this paper we will consider the task of identifying fraudulent online bank account opening requests within a large consumer bank~\citep{jesus2022turning}. In this scenario, fraudsters employ tactics such as identity theft or inventing fictitious individuals to gain access to banking services. A positive prediction (indicating fraud) results in the rejection of the customer's bank account application as a punitive measure, whereas a negative prediction results in the approval of a new bank account and credit card as an assistive measure. Each row in the dataset represents an individual application, all of which were submitted through an online platform with the applicant's explicit consent to store and process their data.

The dataset encompasses eight months of data. The prevalence of fraud fluctuates between 0.85\% and 1.5\% of the instances across different months, with higher values observed in the later months. Furthermore, the distribution of applications also varies from month to month, ranging from 9.5\% on the lower end to 15\% on the higher end.

\begin{itemize}
    \item \textbf{Group Size Disparity}: indicates differences in group frequencies within a dataset, often due to imbalanced populations or uneven adoption of applications among demographics. In particular groups are related to younger (age < 50) or older (50+) individuals;
    \item \textbf{Prevalence Disparity}: emerges when class probabilities rely on the protected group, enabling dataset creation where label probabilities are tied to these groups. It displays higher fraud rates among older age groups, possibly due to fraudsters targeting them for larger bank credit lines;
    \item \textbf{Separability disparity}: goes beyond prior concepts, considering both input features and labels' joint distribution. For instance, in ATM transactions, illumination and age may interact. If the 20-40 age group often uses poorly lit ATMs, it heightens the risk of card cloning. The illumination feature can detect fraud within this group but not others, illustrating separability disparity's intricacies.
\end{itemize}

There is one base dataset and 5 dataset variants with additional patterns. Each variant follows the same underlying distribution as the base dataset but has the following characteristics:
\begin{itemize}
    \item \textbf{Variant I}: Group size disparity worsens, reducing the minority group to 10\%, without prevalence disparities;
    \item \textbf{Variant II}: Steeper prevalence disparities (one group five times the fraud rate), with equal group sizes;
    \item \textbf{Variant III}: Has better separability (fraud/non-fraud) for the the majority group;
    \item \textbf{Variant IV}: Temporal bias introduced with prevalence disparities for the first six months only;
    \item \textbf{Variant V}: Temporal bias with balanced group size and prevalence, featuring feature distribution shifts over time, akin to fraudsters adapting to evade detection.

\end{itemize}
Therefore, variants IV and V present not only difference on prevalence/groups sizes and separability, but also drift with respect to training and testing data.

% \begin{figure*}[t]
%     \centering
%     \begin{tikzpicture}[node distance=2cm, auto]
%         % Upper figure with two boxes and an arrow
%         \node[draw, rectangle, minimum width=2cm, minimum height=1cm] (modelLogic1) {ML Model};
%         \node[draw, rectangle, minimum width=2cm, minimum height=1cm, right=of modelLogic1] (businessLogic1) {Business Logic};
%         \draw[->] (modelLogic1) -- (businessLogic1);
%     \end{tikzpicture}
%     \caption{Common solution in ML adoption.}
%     \label{fig:your_label}
% \end{figure*}
%         % % Space between the two subfigures
%         % \node[below=1cm of modelLogic1] (spacer) {};
% \begin{figure*}[t]
%     \centering
%     \begin{tikzpicture}[node distance=2cm, auto]
%         % Lower figure with three boxes and arrows
%         \node[draw, rectangle, minimum width=2cm, minimum height=1cm] (modelLogic2) {ML Model};
%         \node[draw, rectangle, minimum width=2.5cm, minimum height=1cm, right=of modelLogic2] (scoreDecoupling) {Score Decoupling};
%         \node[draw, rectangle, minimum width=2cm, minimum height=1cm, right=of scoreDecoupling] (businessLogic2) {Business Logic};

%         \draw[->] (modelLogic2) -- (scoreDecoupling);
%         \draw[->] (scoreDecoupling) -- (businessLogic2);
%     \end{tikzpicture}
%     \caption{Decoupled Solution.}
%     \label{fig:your_label}
% \end{figure*}

\section{Calibration Techniques Overview}

We champion calibrating ML models to produce probability estimates that not only represent the true likelihood of an instance belonging to a particular class but also remain consistent during successive retraining, allowing an effective decoupling between machine learning (ML) models and business logic
 
While a myriad of calibration techniques exist for binary classifiers, this paper takes in consideration four pivotal methods: Platt Scaling, Isotonic Regression, Temperature Scaling and Beta Calibration. Some critics might question the decision to limit our discussion to just two techniques amidst a sea of options. However, our rationale is twofold: simplicity and efficacy. We contend that these four methods, in their essence, are representative enough to illuminate the overarching principles of calibration, making our discourse both concise and impactful.

Let $y_i$ for $i = 1, 2, \ldots, n$, be a set of true binary class labels of examples $x_i$, and $f(x_i) = s_i$ the scores for such examples given by a classifier $f(.)$.

\subsection{Platt Scaling}

Platt Scaling~ \citep{platt1999probabilistic} is a post-processing method that converts binary classifiers raw scores into probability estimates. Specifically, given a raw score $s$ produced by a classifier, Platt Scaling maps this score to a probability using logistic regression:

\begin{align*}
P(y=1|x) = \frac{1}{1 + \exp(A \cdot s + B)},
\end{align*}
where $P(y=1|x)$ is the probability of instance $x$ belonging to the positive class (1), $s$ is the output score of the classifier. $ A $ and $ B $ are parameters learned by minimizing the log-likelihood of the logistic model on a validation dataset. 

Then the log-likelihood to be minimized is given by:
\begin{align*}
L(A, B) = -\sum_{i=1}^{n} [y_i \cdot \log\left(\frac{1}{1 + e^{-(A f(x_i) + B)}}\right) + (1 - y_i) \cdot \log\left(1 - \frac{1}{1 + e^{-(A f(x_i) + B)}}\right)],
\end{align*}
where finding $A$ and $B$ that provides the best scaling and shift, respectively for the raw scores, that is further compressed by a logistic function. Note it assumes that a linear combination of the raw scores, followed by a logistic function is sufficient to calibrate the scores. For multiclass scenarios, a one-vs-all approach is commonly used.

\subsection{Isotonic Regression}

Isotonic Regression~\citep{zadrozny2002transforming} is a non-parametric technique used to map classifier scores to probabilities. The idea is to find a set of probabilities $P = \{p_1, p_2, \ldots, p_n\}$ such that $p_i$ is monotonically increasing with $s_i$, while minimizing the weighted mean squared error between the probabilities and true labels. The optimization problem can be formalized as:

$$ L = \min_P \sum_{i=1}^{n} w_i (y_i - p_i)^2 $$

Subject to the constraint that $p_1 \leq p_2 \leq \ldots \leq p_n$, where $y_i$ are the true labels and $w_i$ are instance weights.

\subsection{Temperature Scaling}

Temperature scaling~\citep{guo2017calibration} is a post-processing calibration technique that can be applied to any model producing raw scores or logits. Given scores $ s $ (pre-softmax values or analogous outputs), these scores are scaled by a learned temperature $ T $ before calibrating them. 

In practice, for models like deep neural networks which use softmax we have:
$$P(y=k|s) = \frac{\exp(\frac{s_k}{T})}{\sum_{j}\exp(\frac{s_j}{T})},$$
where $ k $ denotes the class. 

For models producing raw scores, a similar scaling can be applied:
$$ P(y=1|s) = \frac{1}{1 + \exp(\frac{-s}{T})}.$$

In both cases, $ T $ is optimized on a validation set to minimize the discrepancy, such as negative log likelihood, between the predicted probabilities and true labels. A temperature value of 1 implies no modification to the original scores, while $ T > 1 $ softens the probability distribution, approximated probabilities.

\subsection{Beta Calibration}

Beta calibration \citep{Kull2017BetaCA} was introduced as a method to generalize Platt scaling. It models the predicted probability 
$p$ using the following transformation:

\begin{align*}
m(s; a,b,c) = \frac{s^a \cdot c}{s^a \cdot c +(1-s)^b}.
\end{align*}

Here $a,b$, and  $c$ are parameters that are learned from the data. $s$ is the score obtained from the model. The parameters are learnt by minimizing the negative log likelihood $L(a, b, c)$:
\begin{align*}
L(a, b, c) = -\sum_{i=1}^{N} &[ y_i \log(m(s_i; a, b, c)) + (1 - y_i) \log(1 - m(s_i; a, b, c))].
\end{align*}

We can see that Platt scaling is a special case of beta calibration when 
$b=0$ and $c=1$ \citep{Kull2017BetaCA}.

\section{Experimental Design}

\begin{figure}[t]
    \centering
    \includegraphics[width=0.75\linewidth]{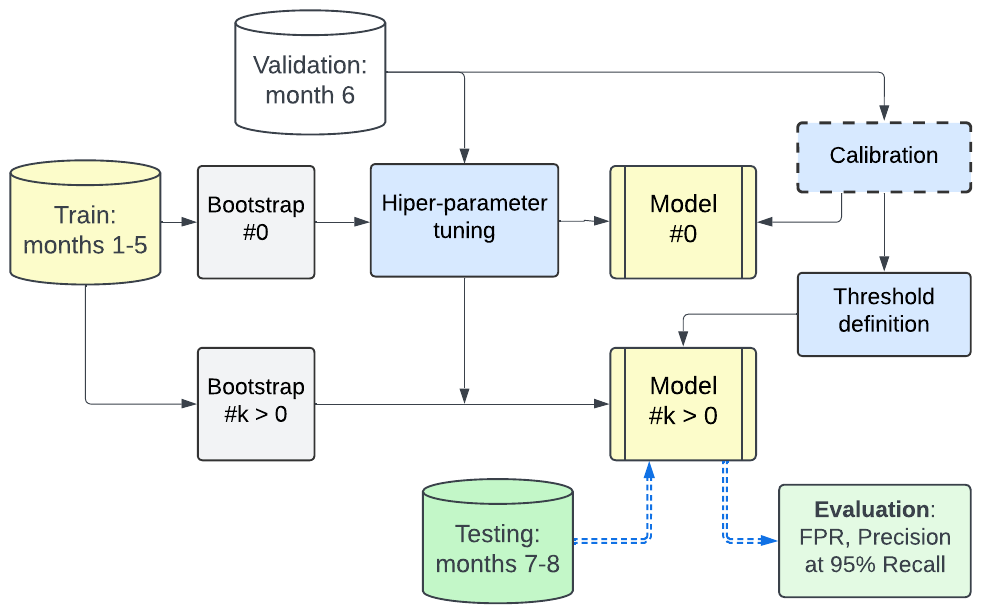}
    \caption{Diagram for the Experimental Setup: obtaining bootstraps of the training set to train several models tuned and calibrated using a validation set, and evaluation using a test set.}
    \label{fig:experiments}
\end{figure}

To provide empirical evidence of decoupling advantages, we orchestrated an experiment using the BAF (Bank Account Fraud) detection dataset~\citep{jesus2022turning} detailed in Section~\ref{sec:baf}. The experiments follow a protocol encompassing the steps below, and illustrated in Figure~\ref{fig:experiments}:
\begin{enumerate}
\item {\bf Bootstrapping:} for each iteration, a bootstrapped sample from the fraud detection training dataset is procured;

\item {\bf Training Phase:} models are trained on the training subset drawn from the current bootstrapped sample. The hyperparameters are found using the first bootstrap as training, and applied to the following bootstraps;

\item {\bf Calibration Phase:} after training, models may be calibrated using as reference the validation subset. We compare a non-calibrated version (skipping this step) and the calibration methods: sigmoid, isotonic, beta and temperature scaling;

\item {\bf Threshold Setting:} in the initial bootstrap iteration (\# 0), we set a decision threshold aiming for a 95\% recall on the validation set. This threshold remains consistent for evaluations in all subsequent bootstrap iterations. Here is where actual decoupling happens.

\item {\bf Evaluation Phase:} the performance of these calibrated models is then gauged on the testing set using Precision at 95\% Recall since it is a metric often used in practical fraud applications, and True Positive Rate at 5\% False Negative Rate, since this is the metric used in the BAF paper~\citep{jesus2022turning}.
\end{enumerate}

\subsection{Statistical test}
The Wilcoxon signed rank test is used to compare the result. It is a non-parametric statistical test comparing the distribution of the signed differences between the paired observations to a null hypothesis distribution of randomly signed differences. We compared the uncalibrated results against the calibrated ones, and considered significant the differences for which the test resulted in $p$-values $\leq 0.01$.

\subsection{Results}
% \begin{center}
% \scalebox{0.75}{
\begin{table*}[t]
\begin{center}
\scalebox{.9}{
\begin{tabular}{llllllll}
\toprule                   &      &       {\bf Base }&  {\bf Variant\_I }& {\bf  Variant\_II }& {\bf Variant\_III }& {\bf Variant\_IV }&  {\bf  Variant\_V } \\
{\bf Model Name } & {\bf  Calibration }
&             &             &             &             &            &            \\
\midrule
{\bf Catboost }& not calibrated &  13.5 ± 1.0 &  12.0 ± 1.0 &  13.3 ± 1.0 &  12.6 ± 1.0 &  8.1 ± 0.4 &  6.4 ± 0.6 \\                   & temperature scaling &  12.4 ± 0.8 &  11.8 ± 0.8 &  12.4 ± 0.6 &  12.5 ± 0.6 &  8.1 ± 0.4 &  6.7 ± 0.4 \\                   & sigmoid &  12.3 ± 0.8 &  11.6 ± 0.8 &  12.3 ± 0.6 &  12.4 ± 0.6 &  8.0 ± 0.4 &  6.6 ± 0.4 \\                   & isotonic & {\bf 13.0 ± 1.8 }&  {\bf 12.2 ± 1.8 }&  {\bf 13.0 ± 1.8 } & {\bf 13.0 ± 2.0 } & {\bf 8.8 ± 1.4 } & {\bf 7.0 ± 0.8 }\\                   & beta &  13.0 ± 1.6 &  11.9 ± 1.0 &  12.4 ± 0.6 &  12.6 ± 0.6 &  8.8 ± 1.2 &  6.7 ± 0.4 \\
\midrule
{\bf Light GBM} & not calibrated & {\bf 13.5 ± 1.0} &  11.8 ± 0.8 &  12.9 ± 0.8 &  12.5 ± 0.8 &  7.9 ± 0.4 &  6.1 ± 0.4 \\                   & temperature scaling &  12.4 ± 0.6 &  11.7 ± 0.6 &  12.1 ± 0.6 &  12.5 ± 0.6 &  7.9 ± 0.4 &  6.4 ± 0.2 \\                   & sigmoid &  12.3 ± 0.6 &  11.6 ± 0.6 &  12.0 ± 0.6 &  12.4 ± 0.6 &  7.8 ± 0.4 &  6.3 ± 0.2 \\                   & isotonic &  13.0 ± 1.4 & {\bf 12.0 ± 1.4 } &  {\bf 12.7 ± 1.8} & {\bf 13.4 ± 1.8 } & {\bf 8.3 ± 1.2} & {\bf 6.7 ± 0.8 }\\                   & beta &  12.4 ± 0.6 &  11.7 ± 0.6 &  12.2 ± 0.6 &  12.7 ± 0.6 &  8.0 ± 0.4 &  6.4 ± 0.2 \\
\midrule
{\bf MLP Neural Network }& not calibrated & {\bf 15.0 ± 3.0 }& {\bf 11.0 ± 4.0 }& {\bf 14.0 ± 5.0} &  11.0 ± 3.0 &  6.4 ± 1.4 &  5.2 ± 1.6 \\                   & temperature scaling &  12.6 ± 1.2 &  11.2 ± 1.4 &  13.0 ± 3.0 & {\bf 12.0 ± 3.0 } &  7.2 ± 1.2 &  {\bf 6.7 ± 1.8 } \\                   & sigmoid &  12.1 ± 1.2 &  10.5 ± 1.4 &  11.4 ± 1.6 &  11.2 ± 1.8 &  6.7 ± 0.8 &  5.8 ± 0.8 \\                   & isotonic &  12.8 ± 1.8 &  10.8 ± 1.8 &  12.0 ± 2.0 &  12.0 ± 2.0 & {\bf 7.3 ± 1.4} &  6.5 ± 1.8 \\                   & beta &  12.4 ± 1.2 &  10.9 ± 1.4 &  12.1 ± 1.8 &  11.7 ± 1.8 &  7.1 ± 1.0 &  6.7 ± 1.4 \\
\bottomrule
\end{tabular}
}
\end{center}
\caption{\it{Precision metric at 95\% Recall. Comparison of the various models and calibration methods in the different dataset variations. Bold values indicate the higher absolute value for each dataset variant, * indicates statistical significance with $p$-value $\leq0.01$ when comparing calibration with non calibrated classifiers.}}
\label{table:table_prec}
\end{table*}
% }
% \end{center}

\begin{table*}[t]
\begin{center}
\scalebox{.9}{
\begin{tabular}{llllllll}
\toprule                   &      &       {\bf  Base} &  {\bf  Variant\_I }&  {\bf  Variant\_II }& {\bf  Variant\_III }& {\bf  Variant\_IV }&  {\bf  Variant\_V} \\
{\bf Model Name} & {\bf Calibration} &              &             &              &              &             &             \\
\midrule
{\bf Catboost}        & not calibrated &   51.9 ± 0.4 &  48.1 ± 0.4 &   52.4 ± 0.6 &   69.2 ± 0.4 &  42.8 ± 0.6 &  32.3 ± 0.4 \\
& temperature scaling &   51.9 ± 0.8 &  48.1 ± 0.8 &   52.4 ± 0.8 &   69.2 ± 0.8 &  42.8 ± 1.2 &  32.3 ± 0.8 \\
& sigmoid             &   51.9 ± 0.8 &  48.1 ± 0.8 &   52.4 ± 0.8 &   69.2 ± 0.8 &  42.8 ± 1.2 &  32.3 ± 0.8 \\
& isotonic            & {\bf 52.4 ± 8.3 }& {\bf 50.6 ± 4.1 }&  {\bf 55.0 ± 3.8 }&  {\bf 71.7*± 3.1} & {\bf 44.7*± 3.0} & {\bf 35.0*± 2.9 }\\
& beta                &   52.0 ± 0.8 &  48.3 ± 1.2 &   53.3 ± 2.0 &   70.0 ± 2.2 &  44.0*± 2.2 &  33.5*± 1.9 \\
\midrule
{\bf Light GBM}     & not calibrated      &   51.5 ± 1.2 &  47.0 ± 0.4 &   50.9 ± 0.4 &   68.3 ± 0.3 &  41.4 ± 0.6 &  31.2 ± 0.5 \\
                    & temperature scaling &   51.6 ± 0.8 &  47.0 ± 0.4 &   51.0 ± 0.4 &   68.3 ± 0.3 &  41.6 ± 0.6 &  31.1 ± 0.5 \\
                    & sigmoid             &   51.6 ± 0.8 &  47.0 ± 0.4 &   51.0 ± 0.4 &   68.3 ± 0.3 &  41.6 ± 0.6 &  31.6 ± 0.5 \\
                    & isotonic            &  {\bf 54.3 ± 4.7 }& {\bf 48.0 ± 3.1 }& {\bf  52.5 ± 3.2 }& {\bf 68.9 ± 9.4 }& {\bf 43.9*± 3.8 }&  {\bf 33.0*± 2.9 }\\
                    & beta                &   52.1 ± 1.8 &  47.0 ± 0.4 &   51.0 ± 0.4 &   68.5 ± 0.7 &  42.0*± 1.7 &  32.4*± 0.9 \\
                    \midrule
{\bf MLP Neural Network} & not calibrated &   46.7 ± 1.2 &  41.6 ± 0.9 &   46.5 ± 0.8 &   65.4 ± 0.8 &  38.4 ± 0.9 &  29.6 ± 0.9 \\
                    & temperature scaling &   47.3 ± 1.0 &  41.4 ± 1.0 &   46.9 ± 1.0 &   66.0 ± 1.0 &  38.9 ± 0.8 &  32.0*± 1.2 \\
                    & sigmoid             &   47.2 ± 1.1 &  42.4 ± 1.0 &   46.8 ± 0.8 &   65.4 ± 0.8 &  38.7 ± 0.9 &  30.0 ± 1.2 \\
                    & isotonic            &  {\bf 49.6 ± 4.1 }& {\bf 43.7 ± 3.5 }& {\bf 48.0 ± 7.6} &  {\bf 67.0 ± 3.0 }& {\bf 42.7*± 4.3 }&  {\bf 33.0*± 3.0} \\
                    & beta                &   47.1 ± 1.2 &  42.7 ± 0.8 &   47.0 ± 1.0 &   66.0 ± 1.8 &  39.7 ± 2.4 &  32.0*± 2.9 \\
\bottomrule
\end{tabular}

}
\end{center}
\caption{\it{True positive rates at 5\% False negative rate. Comparison of the various models and calibration methods in the different dataset variations. Bold values indicate the higher absolute value for each dataset variant, * indicates statistical significance with $p$-value $\leq0.01$ when comparing calibration with non calibrated classifiers.}}
\label{table:table_tpr}
\end{table*}

From the two tables presented, i.e. Table \ref{table:table_prec} and Table \ref{table:table_tpr}, it is evident that models that have been decoupled generally exhibit performance similar to and in some cases better than the not calibrated models. 

The dataset Base, Variants I and II do not have concept drift between training and test sets. In such cases, process of decoupling does not significantly compromise the efficacy of the models in scenarios. For the Variants III, IV and V the methods: isotonic, beta and temperature scaling achieved superior results. This suggests that the isotonic decoupling might offer additional benefits or optimizations for specific types of data or problem contexts, making it a noteworthy approach for those particular scenarios. This observation underscores the importance of evaluating various calibration techniques against different datasets to discern the most effective method for a given application. 

Our primary assumption is unambiguous: we firmly believe that even if we set a threshold specifically to maximize a given metric from the very first iteration, the decoupled models have the capability to not just match, but potentially surpass the performance of models that have not undergone decoupling when it comes to precision or true positive rates on the test datasets. By stating this, we demonstrate a significant point: {\it it is indeed possible to distinguish and separate the functioning of the model from the underlying business processes and logic without compromising safety or consistency.}

\section{Conclusion}

In this paper, we addressed the critical question of the separation between machine learning logic and business logic. By meticulously setting up controlled experiments, our findings not only confirm the intuitive benefits of such a separation, but also shed light on some non-trivial advantages.

Our results underscore the importance of probability calibration in maintaining a clear boundary between machine learning logic, and business logic. Such separation fosters improved maintainability by enabling independent modifications and upgrades, thereby ensuring that updates to the business logic do not inadvertently affect the ML model, and vice versa.

Furthermore, scalability greatly benefits from this approach. Since each model can be independently used based on its own skill, it reduces the potential bottleneck situations where we need different logic for different use cases. Stability also witnesses a marked improvement, as any instability or bugs in one domain remain isolated, preventing cascading failures across the system.

Interestingly, while the aforementioned benefits center around system design principles, our experiments also brought to the fore an empirical observation. In some cases, the very act of separating the two logics via probability calibration not only maintained, but in fact, enhanced the model’s performance. This indicates that the separation might allow each logic to focus on its core functionality, thereby optimizing its operation.

In conclusion, our research makes a compelling case for the clear demarcation between machine learning logic and business logic using probability calibration. Not only does this methodology pave the way for more robust, maintainable, and scalable systems, but it also has the potential to yield performance dividends. As the realms of machine learning and business operations continue to converge, it is our belief that such an architectural separation will become a gold standard in system design.

\section*{Acknowledgments}
  The authors would like to thank Pedro Agustin Carossi and Mercado Libre Inc. for their support in this research.

\bibliographystyle{unsrtnat}
\bibliography{sample-bibliography} 
% Data statement is mandatory for all papers
\appendix

\section{Appendix}
\subsection{Data}
As dataset for out experiment we used Bank Account Fraud Dataset Suite (NeurIPS 2022), introduced at NeurIPS 2022. It encompasses a series of six distinct synthetic tabular datasets oriented towards bank account fraud detection. Unveiled as a pioneering endeavor in its realm, this suite seeks to provide a comprehensive and robust platform for the assessment of both emergent and established machine learning (ML) techniques, including those focusing on fairness in ML.

Characteristics of the BAF dataset suite are delineated as follows:
\begin{enumerate}

\item Realism: Derived from contemporary real-world data pertinent to fraud detection, the BAF datasets emulate the intricacies and nuances inherent in genuine transactional records.

\item Controlled Bias: Each dataset within the suite is specifically crafted to exhibit unique, predefined types of bias. This is vital for testing the resilience and adaptability of ML algorithms under varying biased conditions.

\item Class Imbalance: Mirroring the rarity of fraudulent activities in actual transactional contexts, the datasets are marked by an exceptionally low prevalence of the positive (fraudulent) class. This characteristic introduces the challenge of distinguishing rare events from a sea of legitimate transactions.

\item Temporal Dynamics: The datasets are not only chronologically structured but also exhibit observed distribution shifts over time. This facet ensures that algorithms are tested for their ability to adapt and remain efficacious in evolving scenarios.

\item Privacy Considerations: Given the paramount importance of individual privacy, especially in financial contexts, a multipronged approach has been adopted to safeguard the identity of potential account holders. This entails the integration of differential privacy measures (via noise infusion), feature encryption, and the deployment of a generative model known as CTGAN. These strategies collectively obfuscate sensitive details while preserving the overarching data structure and relationships.

\end{enumerate}

The BAF dataset suite stands as a meticulously designed tool for the scientific community, bridging the gap between theoretical algorithmic advancements and their practical applications in a sensitive domain like fraud detection.

\subsection{Practical Examples of Decoupling Decision Making from Model Outputs}
\label{app:examples}
\begin{itemize}

\item Fraud Detection in Financial Transactions: In the financial sector, ML models are extensively used to identify potentially fraudulent activities and suspicious transactions. Decoupling the fraud detection model from predefined intervention thresholds is pivotal. This ensures that decisions related to flagging and investigating potential frauds remain consistent and are driven by factors independent of the model's performance, such as regulatory guidelines, risk appetites of the institution, and historical fraud patterns. 

\item Healthcare Patient Risk Stratification: In healthcare, ML models can help predict patient outcomes, such as the likelihood of readmission or complications after surgery. Decoupling the patient risk stratification model from the predefined risk thresholds allows healthcare providers to make consistent treatment decisions based on factors independent of the model's performance, such as clinical guidelines, resource allocation, and patient preferences.

\item Customer Churn Prediction and Retention Strategy: ML models can help businesses predict which customers are likely to churn or cancel their services. Decoupling the churn prediction model from predefined intervention thresholds is essential, as it ensures that customer retention strategies remain consistent and are driven by factors independent of the model's performance, such as business objectives, budget constraints, and customer segmentation.

\item Anomaly Detection in Industrial Processes: In manufacturing and industrial settings, ML models can be employed to monitor equipment, production lines, or processes to identify anomalies that may indicate potential issues, such as equipment failure, quality issues, or safety hazards. Decoupling the anomaly detection model from predefined intervention thresholds is critical, as it ensures that maintenance and safety decisions remain consistent and are driven by factors independent of the model's performance, such as company policies, safety regulations, and resource availability.

\item Marketing Campaign Optimization: In the marketing domain, ML models can help predict customer responses to different marketing campaigns, enabling businesses to optimize their marketing efforts and maximize return on investment. Decoupling the marketing campaign optimization model from predefined decision thresholds is essential, as it ensures that marketing decisions remain consistent and are driven by factors independent of the model's performance, such as business objectives, budget constraints, and competitive landscape.

\end{itemize}

These examples highlight the importance of decoupling techniques in real-world applications, demonstrating their value in creating efficient and adaptable ML-driven systems.

\end{document}